\documentclass[runningheads]{llncs}
\usepackage[T1]{fontenc}
\usepackage{graphicx}
\usepackage{cite}
\usepackage{amsmath,amssymb,amsfonts}
\usepackage{graphicx}
\usepackage{textcomp}
\usepackage{booktabs}
\usepackage{xcolor}
\usepackage{algorithm}
\usepackage{algorithmicx}
\usepackage{algpseudocode}
\usepackage{multirow}
\usepackage{caption} 
\usepackage{subcaption}
\usepackage{hyperref}
\usepackage{url}
\usepackage{fontawesome5}

\begin{document}
\title{Catastrophic Forgetting Mitigation via Discrepancy-Weighted Experience Replay\thanks{Supported by the National Key R\&D Program of China under Grant 2022YFF0503900, and the Research Project of Institute of Software, Chinese Academy of Sciences (ISCAS-ZD-202401, ISCAS-ZD-202403).}}

\author{
Xinrun Xu\inst{1, 2, \dag}  \and
Jianwen Yang\inst{1, 2, \dag}  \and
Qiuhong Zhang\inst{1, 2} \and
Zhanbiao Lian\inst{1, 2} \and \\
Zhiming Ding\inst{1, \textsuperscript{\faEnvelope}} \and 
Shan Jiang\inst{3, \textsuperscript{\faEnvelope}} 
}

\authorrunning{X. Xu, J. Yang et al.}
\titlerunning{ER-EMU}

\institute{Institute of Software Chinese Academy of Science, Beijing, China \and
University of Chinese Academy of Sciences, Beijing, China \and
Advanced Institute of Big Data, Beijing, China
}

\maketitle 
\begingroup\renewcommand{\thefootnote}{\textsuperscript{\dag}}
\footnotetext{Equal Contribution.}
\endgroup

\begin{abstract}
Continually adapting edge models in cloud-edge collaborative object detection for traffic monitoring suffers from catastrophic forgetting, where models lose previously learned knowledge when adapting to new data distributions. This is especially problematic in dynamic traffic environments characterised by periodic variations (e.g., day/night, peak hours), where past knowledge remains valuable. Existing approaches like experience replay and visual prompts offer some mitigation, but struggle to effectively prioritize and leverage historical data for optimal knowledge retention and adaptation. Specifically, simply storing and replaying all historical data can be inefficient, while treating all historical experiences as equally important overlooks their varying relevance to the current domain. This paper proposes ER-EMU, an edge model update algorithm based on adaptive experience replay, to address these limitations. ER-EMU utilizes a limited-size experience buffer managed using a First-In-First-Out (FIFO) principle, and a novel Domain Distance Metric-based Experience Selection (DDM-ES) algorithm. DDM-ES employs the multi-kernel maximum mean discrepancy (MK-MMD) to quantify the dissimilarity between target domains, prioritizing the selection of historical data that is most dissimilar to the current target domain. This ensures training diversity and facilitates the retention of knowledge from a wider range of past experiences, while also preventing overfitting to the new domain. The experience buffer is also updated using a simple random sampling strategy to maintain a balanced representation of previous domains. Experiments on the Bellevue traffic video dataset, involving repeated day/night cycles, demonstrate that ER-EMU consistently improves the performance of several state-of-the-art cloud-edge collaborative object detection frameworks. Specifically, methods enhanced by ER-EMU show improved adaptation to repeated scenarios and significant performance gains over baselines, highlighting its effectiveness in mitigating catastrophic forgetting, improving adaptability, and enhancing model generalization in dynamic traffic monitoring scenarios. Furthermore, ablation studies show that DDM-ES plays a crucial role in the effectiveness of ER-EMU compared to a random historical experience selection, and the influence of the parameter $l$ on the ER-EMU algorithm is also experimentally analyzed.
\end{abstract}

\section{Introduction}

Continuously retraining edge models in cloud-edge-end collaborative architectures presents a significant challenge due to catastrophic forgetting \cite{khani2021real,ding2020cloud,li2019rilod,nan2023large,ding2022internet,ding2022emergency}. As these models adapt to new data distributions, they risk losing previously learned knowledge \cite{mccloskey1989catastrophic}, potentially leading to inconsistent performance and difficulties in handling recurring scenarios \cite{gan2023cloud,xu2024multi}. This is particularly critical in dynamic environments like traffic monitoring, where conditions such as day/night transitions and peak hour traffic patterns exhibit cyclical behavior, making previously learned knowledge highly valuable. The inability to retain this past knowledge impedes the model's ability to effectively adapt and generalize across these repeating but varying scenarios.

\begin{figure*}[!h]
    \centering
    \includegraphics[width=\linewidth]{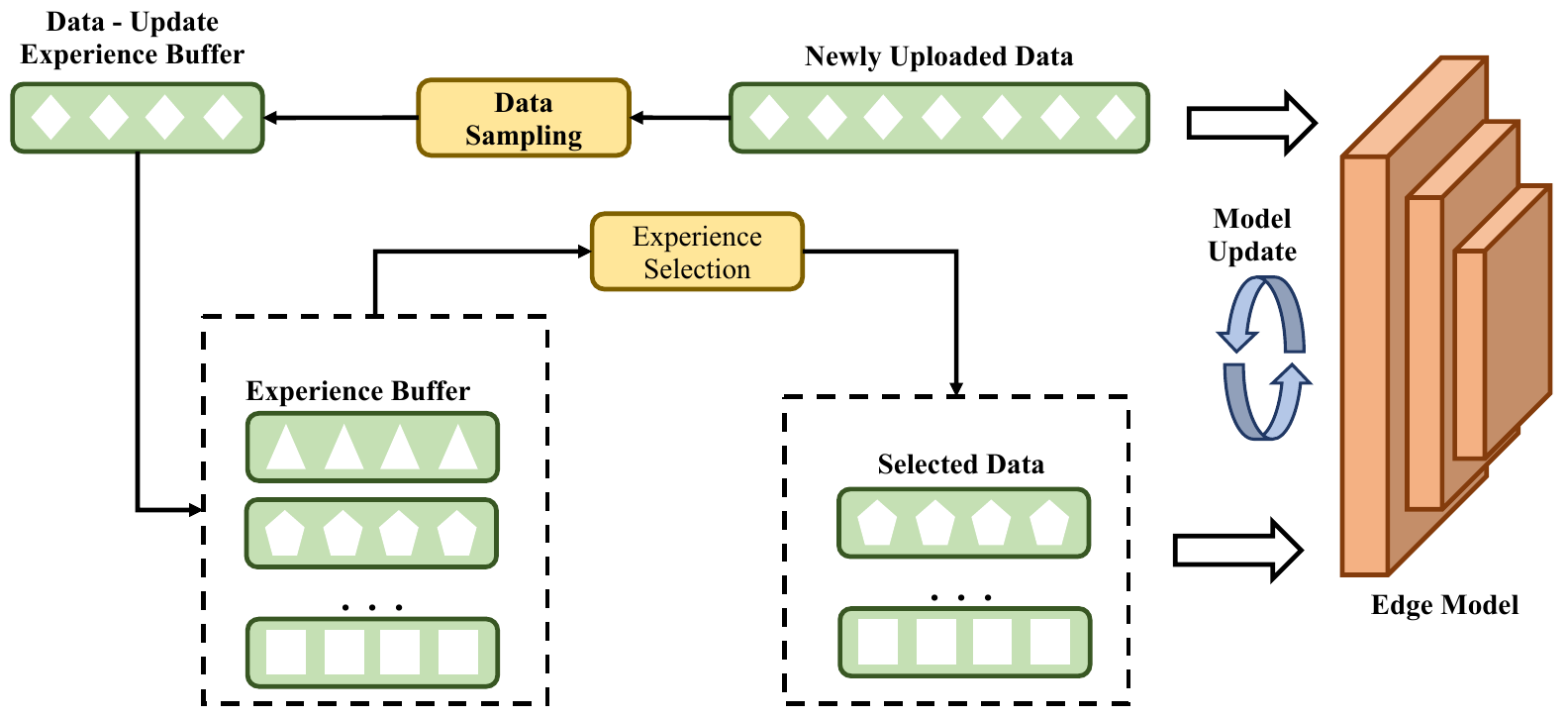}
    \caption{Illustration of the edge model update process based on adaptive experience replay. When the edge model receives new data from the current target domain, the ER-EMU algorithm selects historical data from the experience buffer for training based on domain dissimilarity. 
    }
    \label{fig:经验回放}
\end{figure*}

Retraining from scratch with all available data, while a direct solution \cite{parisi2019continual,rebuffi2017icarl}, is often infeasible in resource-constrained edge environments. This approach also risks overfitting and may hinder the model's ability to generalize effectively, as the sheer volume of data could overwhelm the edge model’s capacity for broad learning. Furthermore, continuously retraining from scratch significantly undermines the timeliness of updates in these dynamic and time-sensitive environments. Recent efforts like Shoggoth \cite{wang2023shoggoth} employ experience replay with adaptive mini-batch training to alleviate catastrophic forgetting. While this represents a step forward, such methods often lack a mechanism for discerning the relative importance of past experiences during retraining. Specifically, these methods often treat all historical data as equally important, failing to recognize that certain past experiences may be more relevant to the current adaptation task than others, and this could negatively impact the efficiency and effectiveness of the retraining process.

This paper proposes a novel edge model update algorithm based on adaptive experience replay, named ER-EMU, to address this limitation. Our algorithm leverages a tailored experience cache and management mechanism to selectively incorporate pertinent historical data during retraining. Unlike previous approaches, ER-EMU dynamically prioritizes past experiences using a domain distance metric, thus focusing on knowledge that is most relevant for current adaptation and thus preventing overfit to the current domain. This approach aims to mitigate catastrophic forgetting effectively, preserve historical knowledge, and improve the model's adaptability while remaining sensitive to the dynamic nature of edge environments \cite{lian2024cloud,jiang2021graph,xu2025high}. The proposed method leverages both the recurring nature of traffic patterns and selectively leverages historic data to improve the performance of edge models.

We begin by formally defining the edge model update problem within the context of cloud-edge-end collaborations, focusing on the challenge of catastrophic forgetting in the context of continuous adaptation. Subsequently, we detail our proposed algorithm and its core components: the experience buffer, the Domain Distance Metric-based Experience Selection (DDM-ES) algorithm, and the random sampling based buffer update approach. Finally, we validate the efficacy of our method through extensive experiments conducted on real-world traffic video datasets, comparing its performance against state-of-the-art approaches and demonstrating the benefits of our adaptive replay mechanism.

\section{Related Works}


In cloud-edge collaborative frameworks, updating edge models to adapt to new environments is crucial but is plagued by catastrophic forgetting \cite{mccloskey1989catastrophic, parisi2019continual}. As edge models continually adapt, they risk losing knowledge acquired from previous domains. This is especially problematic in traffic monitoring due to its cyclical patterns (e.g., day/night cycles), making historical knowledge valuable. The limitations of simply retraining from scratch with all data are resource constraints at the edge, possible overfitting, and impaired timeliness \cite{rebuffi2017icarl}.

Existing work attempts to address these issues. The Shoggoth method \cite{wang2023shoggoth} mitigates catastrophic forgetting through experience replay with a simple random sampling approach, and utilizes adaptive mini-batch training. Other methods, such as \cite{lopez2017gradient} propose gradient episodic memory to better preserve information from past experiences. Aljundi et al. \cite{aljundi2018memory} approach this problem with a method that aims to learn what not to forget. However, these approaches do not distinguish the relative importance of different historical data points during the update. These methods treat all experiences equally, failing to identify when certain past data may be more pertinent for the current adaptation.  DCC \cite{gan2023cloud} addresses catastrophic forgetting using current data with unique visual prompts.  While this has been shown to improve model generalization, it does not incorporate historical experiences for additional improvements.

This paper argues that strategically incorporating relevant historical information from different target domains is key to continual learning, which is particularly true in systems that display periodic changes in traffic patterns. The proposed ER-EMU method addresses these issues by adaptively selecting from historical data, ensuring a comprehensive learning experience while mitigating the risk of overfitting to new target domains.

\section{Formal Description of the Edge Model Update Problem}

This paper addresses the challenge of catastrophic forgetting in the context of a cloud-edge collaborative adaptive object detection framework. Our goal is to enable the lightweight edge model, denoted as $\theta_{edge}$, to maintain its performance on a continuously evolving sequence of target domains, ${D_{t1}, D_{t2}, ..., D_{tT}}$, without sacrificing previously acquired knowledge. Each target domain, $D_{ti} = {(x_j^i,y_j^i)}{j=1}^{N{ti}}$, represents the data distribution encountered during the $i$-th model update cycle, corresponding to traffic monitoring video from a specific location and time period. Here, $N_{ti}$ represents the number of samples in $D_{ti}$, while $x_j^i$ and $y_j^i$ denote the $j$-th sample and its corresponding label, respectively. The labels are generated by the cloud model, $\theta_{cloud}$.

The model learning paradigm involves leveraging knowledge acquired from the previous $i-1$ target domains to effectively adapt to the $i$-th target domain. Notably, the data distribution of each target domain can be arbitrary, although exhibiting some degree of periodicity in its evolution. The objective is formalized as shown in Equation \ref{eq:edge-model-object-function},

\begin{equation}
\label{eq:edge-model-object-function}
\sum_{i=1}^{T} \min \mathcal{L}(\theta_{edge}, D_{ti}) + \lambda \Phi(\theta_{edge}, \{D_{t1}, ..., D_{t(i-1)}\})
\end{equation}
where $\mathcal{L}$ is the loss function for the $i$-th target domain $D_{ti}$, $\Phi$ is the regularization term to prevent catastrophic forgetting, and $\lambda$ is the coefficient balancing their importance. The loss function $\mathcal{L}$ can be any suitable metric for object detection, such as mAP.

This study addresses the challenge of continual learning in cloud-edge collaborative object detection for traffic monitoring applications. Specifically, we aim to develop an effective edge model updating strategy that mitigates catastrophic forgetting while ensuring adaptability to evolving target domains and generalization across both historical and emerging data.

\section{Edge Model Update Algorithm Based on Adaptive Experience Replay}

ER-EMU, an edge model update algorithm that mitigates catastrophic forgetting during adaptation, leverages an experience replay mechanism (Figure \ref{fig:经验回放}). It maintains a FIFO experience buffer storing representative data from past target domains, used strategically during model fine-tuning. When adapting to a new target domain, the algorithm trains on a batch mixing data from the latest domain and the experience buffer. A dedicated sampling algorithm (Section \ref{sec:缓冲区更新}) ensures the buffer's representativeness by selecting which new data points are incorporated.

\subsection{Experience Selection Algorithm Based on Inter-domain Distance Metric} \label{sec:经验选择}

Experience replay, a classic approach to alleviate catastrophic forgetting \cite{aljundi2018memory,lopez2017gradient}, leverages historical data or features during training. Its effectiveness depends on selecting relevant and diverse historical experiences (add citation). In our adaptive object detection scenario, we propose an experience selection algorithm based on inter-domain distance metrics. At each training round, the algorithm calculates the distance between the new target domain and all historical domains in the experience buffer. The $m$ most distant historical domains, determined by this metric, augment the current target domain data during fine-tuning, promoting knowledge preservation from diverse domains. Among various distance metrics, Maximum Mean Discrepancy (MMD) \cite{oza2023unsupervised} is widely used to quantify domain dissimilarity. Given domains $D_a$ and $D_b$, represented by datasets \(X_a=[x_{a1},x_{a2},...,X_{an_a}]\), and \(X_b=[x_{b1},x_{b2},...,x_{bn_b}]\), their MMD is computed using Equation \ref{eq:mmd},

\begin{equation} \label{eq:mmd}
\scriptsize
\begin{aligned}
    H_{MMD} =& Dist_{MMD}(X_a, X_b, k) \\
    =&\left\| \frac{1}{n_a} \sum_{i=1}^{n_a} \phi(x_{i}) - \frac{1}{n_b} \sum_{j=1}^{n_b} \phi(x_{bj}) \right\|^2_{\mathcal{H}} \\
    =& \frac{1}{n_a^2} \sum_{i=1}^{n_a} \sum_{i'=1}^{n_a} \langle \phi(x_{ai}), \phi(x_{ai'}) \rangle_{\mathcal{H}} - \frac{2}{n_a n_b} \sum_{i=1}^{n_a} \sum_{j=1}^{n_b} \langle \phi(x_{ai}), \phi(x_{bj}) \rangle_{\mathcal{H}} \quad \\
    & + \frac{1}{n_b^2} \sum_{j=1}^{n_b} \sum_{j'=1}^{n_b} \langle \phi(x_{bj}), \phi(x_{bj'}) \rangle_{\mathcal{H}} \\
    =& \frac{1}{n_a^2} \sum_{i=1}^{n_a} \sum_{i'=1}^{n_a} k(x_{ai}, x_{ai'}) - \frac{2}{n_a n_b}
    \sum_{i=1}^{n_a} \sum_{j=1}^{n_b} k(x_{ai}, x_{bj}) \quad + \frac{1}{n_b^2} \sum_{j=1}^{n_b} \sum_{j'=1}^{n_b} k(x_{bj}, x_{bj'}) \\
\end{aligned}
\end{equation}
where \(\phi(\cdot)\) is a mapping function that maps the dataset \(X\) to a Hilbert space \(H\) (this step is usually achieved by using a specific kernel function $k(\cdot)$), and \(n_a\) and \(n_b\) represent the number of samples in the two domains, respectively.

However, MMD's reliance on a task-specific kernel function limits its generalizability. To address this, our approach utilizes Multi-Kernel Maximum Mean Discrepancy (MK-MMD) \cite{gretton2012kernel,cortes2012algorithms} as the distance metric. MK-MMD combines multiple kernel functions (Equation \ref{eq:mkmmd}) to represent the discrepancy between distributions, enhancing expressiveness and capturing more complex inter-domain relationships compared to traditional MMD. Consequently, MK-MMD has shown superior performance in cross-domain learning and pattern recognition tasks \cite{gretton2012kernel,cortes2012algorithms}.

\begin{equation} \label{eq:mkmmd}
\begin{aligned}
    H_{MK} = \text{Dist}_{MK}(\mathcal{D}_a, \mathcal{D}_b, \mathcal{K}) = \sum_{u=1}^m \beta_u Dist_{MMD}(X_a, X_b, k_u)
\end{aligned}
\end{equation}
where $\mathcal{K}$ is the set of kernel functions, $m$ is the number of kernel functions, $\beta_u$ is the weight of the $u$-th kernel function $k_u(\cdot)$, satisfying $\sum_{u=1}^d \beta_u = 1 (\beta_u \geq 0)$, and the specific form of $\mathcal{K}$ is shown in Equation \ref{eq:mkmmd-kernels}.

\begin{equation} \label{eq:mkmmd-kernels}
\small
    \mathcal{K}:=\left\{k: k=\sum_{u=1}^{m} \beta_{u} k_{u}, \sum_{u=1}^{m} \beta_{u}=D, \beta_{u} \geq 0, \forall u \in\{1, \ldots, m\}\right\}
\end{equation}

\begin{equation} \label{eq:dti-loss}
\small
\begin{aligned}
    \mathcal{L}_{Det}(\theta_{edge}) &= \mathbf{L}_{t}(\theta_{edge}, D_ti) + \mathcal{L}_{replay}(\theta_{edge}, D_{H}, W_{H}) \\
    &= \mathcal{L}_{t}(\theta_{edge}, D_ti) + \sum_{j=1}^H w_{hj} \mathcal{L}_{t}(\theta_{edge}, D_{hj})
\end{aligned}
\end{equation}

Leveraging MK-MMD, we develop Domain Distance Metric-based Experience Selection (DDM-ES), detailed in Algorithm \ref{alg:DDM-ES}.  If the experience buffer contains historical data, the algorithm computes the MK-MMD between each historical target domain and the current target domain, quantifying their dissimilarity. The $l$ most distant historical domains form a set $D_H$. Each domain in $D_H$ is then weighted using a Sigmoid function parameterized by their distances, ensuring that more dissimilar domains exert greater influence during training. This adaptive weighting in DDM-ES balances the incorporation of past experiences with mitigating overfitting to the current target domain.

\begin{algorithm}
\caption{Domain Distance Metric-based Experience Selection (DDM-ES)}
\label{alg:DDM-ES}
\scriptsize
\begin{algorithmic}[1]
\Statex \hspace*{-\algorithmicindent} \textbf{Input:} Experience buffer $M$, target domain dataset $D_{ti}$.
\Statex \hspace*{-\algorithmicindent} \textbf{Output:} Selected $l$ historical target domain datasets $D_{H} = \{D_{h1}, D_{h2}, ... , D_{hl}\}$, and the weight of each historical target domain dataset $W_{H} = \{w_{h1}, w_{h2}, ... , w_{hl}\}$.
\If{IsEmpty(M)}
    \State \Return
\Else
    \State Initialize the output historical target domain set $D_{H} \gets [\ ]$
    
    \State Initialize the output weight set for historical target domains $W_{H} \gets [\ ]$
    
    \State Initialize the MK-MMD set $dists$ between each historical target domain stored in $M$ and $D_{ti} \gets [\ ]$
    
    \For{ each historical target domain ${D_{hj}}$ stored in $M$}
        \State Calculate the MK-MMD $Dist_{MK}(D_{ti}, D_{hj})$ between $D_{hj}$ and $D_{ti}$ using Equation \ref{eq:mkmmd}
        \State Update $dists$, $dists[j] \gets Dist_{MK}(D_{ti}, D_{hj})$
    \EndFor   
    \State Copy $M$, $M_{copy} \gets M$ 
    \State Sort $M_{copy}$ and $dists$ in ascending order based on $dists$, ($dists$,$M_{copy}$) $\gets sort(M_{copy}, dists)$
    \State Update $D_{H} \gets M_{copy}[:l]$
    \For{ each historical target domain ${D_j}$ in $D_{H}$}
        \State Update the corresponding weight $W_{H}[j] \gets {sigmoid(dists[j])}$
    \EndFor
\EndIf
\end{algorithmic}
\end{algorithm}

\subsection{Experience Buffer Update Algorithm Based on Random Sampling} \label{sec:缓冲区更新}

To maximize data diversity, our approach enforces balanced representation within the experience buffer, storing an equal number of samples from each target domain. Assuming independent and identically distributed data within each domain, a random sampling strategy selects data points for inclusion. This facilitates equitable knowledge sharing across domains (Algorithm \ref{alg:replay_memory}).  When processing a new target domain, if the buffer is full, random samples are removed to create space. Then, random samples from the new domain are incorporated, ensuring the buffer's capacity limit and maximizing knowledge retention from all encountered domains.

\begin{algorithm}
\caption{Random Sampling-based Experience Buffer Update Algorithm (RS-EBU)}
\label{alg:replay_memory}
\scriptsize
\begin{algorithmic}[1]
\Statex \hspace*{-\algorithmicindent} \textbf{Input:} Experience buffer $M$, the number of images $h$ to be saved per domain, target domain dataset $D_{ti}$
\Statex \hspace*{-\algorithmicindent} \textbf{Output:} None
\If{IsFull($M$)} 
    \State Randomly sample $h$ images $M_{\text{add}}$ from $D_{ti}$
    \State Select the oldest $h$ images $M_{\text{replace}}$ from $M$
    \State Update $M \gets (M - M_{\text{replace}}) \cup M_{\text{add}}$
\Else
    \State Randomly sample $h$ images $M_{\text{add}}$ from $D_{ti}$ 
    \State Update $M \gets M \cup M_{\text{add}}$ 
\EndIf
\end{algorithmic}
\end{algorithm}

\subsection{Steps of the Edge Model Update Algorithm Based on Adaptive Experience Replay}

Integrating the DDM-ES and RS-EBU algorithms, we propose the Edge Model Update Algorithm Based on Adaptive Experience Replay (ER-EMU). Upon arrival of a new target domain, $D_{ti}$, DDM-ES identifies a subset of $l$ most dissimilar historical target domain datasets, $D_{H}={D_{h1}, D_{h2}, ... , D_{hl}}$, and computes their corresponding weight set $W_{H} = {w_{h1}, w_{h2}, ... , w_{hl}}$, reflecting their training importance. The loss function for this round of edge model updates is then formulated as shown in Equation \ref{eq:dti-loss}.

After the edge model updates using the combined data, RS-EBU incorporates the new target domain, $D_{ti}$, into the experience buffer, ensuring its representativeness.  ER-EMU optimizes the objective function (Equation \ref{eq:edge-model-object-function}) by strategically recalling relevant past knowledge during each model adaptation cycle, mitigating catastrophic forgetting and enabling the edge model to maintain performance on previously learned domains while adapting to new ones.

\section{Experimental Validation}

The experimental evaluation validates the effectiveness of ER-EMU within a cloud-edge collaborative architecture. Our methodology integrates ER-EMU into several SOTA cloud-edge collaborative object detection frameworks, replacing their native edge model update modules. By comparing performance on real-world traffic monitoring data before and after incorporating ER-EMU, we assess its efficacy in mitigating catastrophic forgetting and enhancing adaptability.

\subsection{Experimental Setup}

Our experimental setup uses a pre-trained DINO \cite{zhang2022dino} model (ResNet-101 \cite{he2015deepresiduallearningimage} backbone) for the cloud server and a lighter RT-DETR model \cite{zhao2024detrsbeatyolosrealtime} (ResNet-18 backbone) for the edge server. The Bellevue traffic video dataset provides real-world traffic scenarios for evaluation. Table \ref{tab:er-emu-parameters} summarizes the hyperparameters used in our experiments.

\begin{table*}[htbp]
\centering
\setlength{\tabcolsep}{6pt}
\renewcommand{\arraystretch}{1.2}
\caption{Experimental hyperparameters for performance validation of ER-EMU algorithm.}
\label{tab:er-emu-parameters}
\resizebox{0.8\linewidth}{!}{
\begin{tabular}{@{} ccc @{}}
\toprule
Hyperparameter & Setting & Description \\
\midrule
$M_{size}$ & 30 & Maximum buffer size for historical target domains. \\
$h$ & 200 & Samples per historical domain. \\
$l$ & 5 & Historical domains selected per update. \\
\bottomrule
\end{tabular}
}
\end{table*}

\textbf{Baselines:}
\textbf{1) Shoggoth \cite{wang2023shoggoth}:} Cloud-edge, random historical data sampling, combats catastrophic forgetting. \textbf{2) Shoggoth-ER-EMU:} Shoggoth enhanced with ER-EMU edge model updates. \textbf{3) DCC \cite{gan2023cloud}:} Cloud-edge, current domain data, visual prompts for forgetting mitigation. \textbf{4) DCC-ER-EMU:} DCC enhanced with ER-EMU edge model updates. \textbf{5) LVACCL:} Cloud-edge, video stream correlations, edge model retraining. \textbf{6) LVACCL-ER-EMU: } LVACCL with ER-EMU, modified temporal correlation handling.

\subsection{Effectiveness of the ER-EMU Algorithm in Mitigating Catastrophic Forgetting in Edge Models}

\begin{table*}[!h]
\caption{Continual Adaptation capability on Bellevue Traffic with repeat scene for ER-EMU.}
\centering
\label{tab:er-emu-repeat-bellevue}
\resizebox{0.8\linewidth}{!}{
\begin{tabular}{lccccccccc} 
\hline
Time & \multicolumn{9}{c}{t$\xrightarrow{\rule{0.5\linewidth}{0pt}}$} \\
\hline
Video & day1 & day2 & night1 & night2 & day1 & day2 & night1 & night2 & Mean  \\
\hline
Shoggoth & 61.9 & 59.7 & 57.5 & 58.6 & 62.4 & 60.3 & 58.9 & 58.9 & 59.8 \\
Shoggoth-ER-EMU & 63.1 & 61.3 & 59.7 & 60.2 & 64.7 & 63.8 & 62.8 & 62.2 & 62.2 \\
DCC & 60.7 & 60.5 & 56.7 & 57.3 & 62.7 & 59.4 & 57.7 & 55.4 & 58.9 \\
{DCC-ER-EMU} & 63.9 & 63.5 & 62.3 & 63.1 & 65.5 & 65.3 & 62.5 & 63.1 & 63.7 \\
LVCCL& 62.5 & 59.4 & 56.3 & 57.2 & 61.5 & 61.3 & 56.9 & 58.7 & 59.2  \\
LVCCL-ER-EMU & 63.6 & 60.0 & 57.0 & 58.2 & 64.2 & 61.7 & 58.2 & 59.1 & 60.3  \\
\hline
\end{tabular}
}
\end{table*}

To validate ER-EMU's effectiveness in addressing catastrophic forgetting, we tested it in an extreme scenario: cycling through two daytime/nighttime Bellevue traffic video scenes twice. ER-EMU was applied to Shoggoth, DCC, and LVACCL. Table \ref{tab:er-emu-repeat-bellevue} shows average mAP for each method per scene and overall. Results indicate that ER-EMU consistently improves detection accuracy for SOTA cloud-edge adaptive methods. Methods using ER-EMU show improved performance upon re-encountering scenes, demonstrating its effectiveness in mitigating catastrophic forgetting. While Shoggoth and DCC (mini-batch experience replay \& visual prompts) show some resistance, ER-EMU further improves their performance, indicating its superiority over the comparison methods' edge model update modules in mitigating catastrophic forgetting.

\subsection{Performance Validation Experiments for the ER-EMU Algorithm}

\begin{figure*}[ht]
\centering
\begin{subfigure}[b]{0.24\linewidth}
    \includegraphics[width=\linewidth]{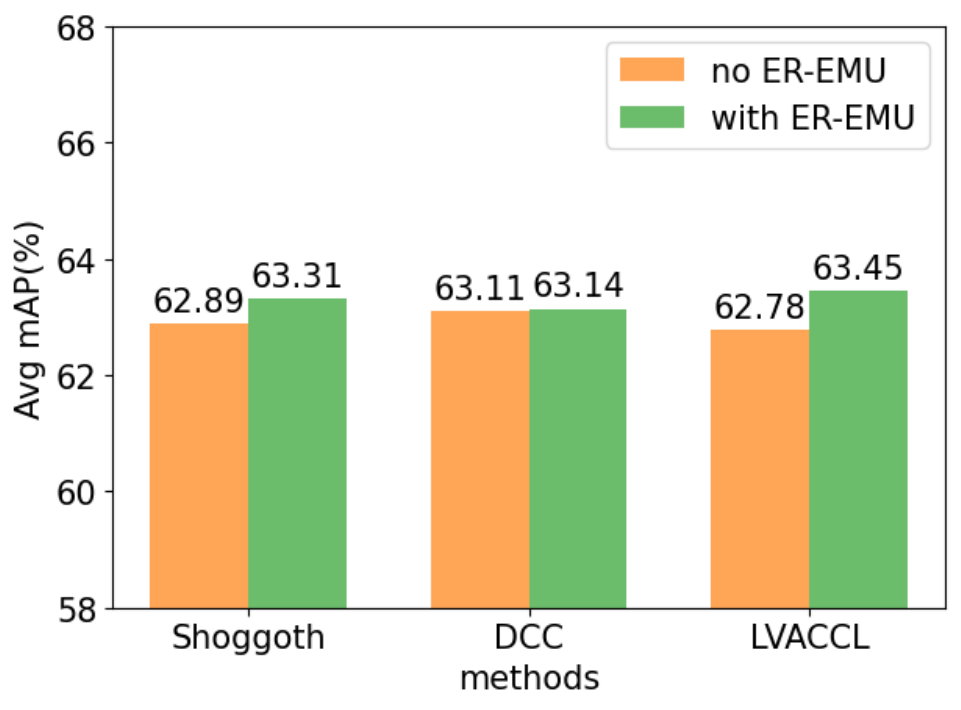}
    \caption{Time Span: 1D}
    \label{fig:经验回放性能对比-sub1}
\end{subfigure}
\begin{subfigure}[b]{0.24\linewidth}
    \includegraphics[width=\linewidth]{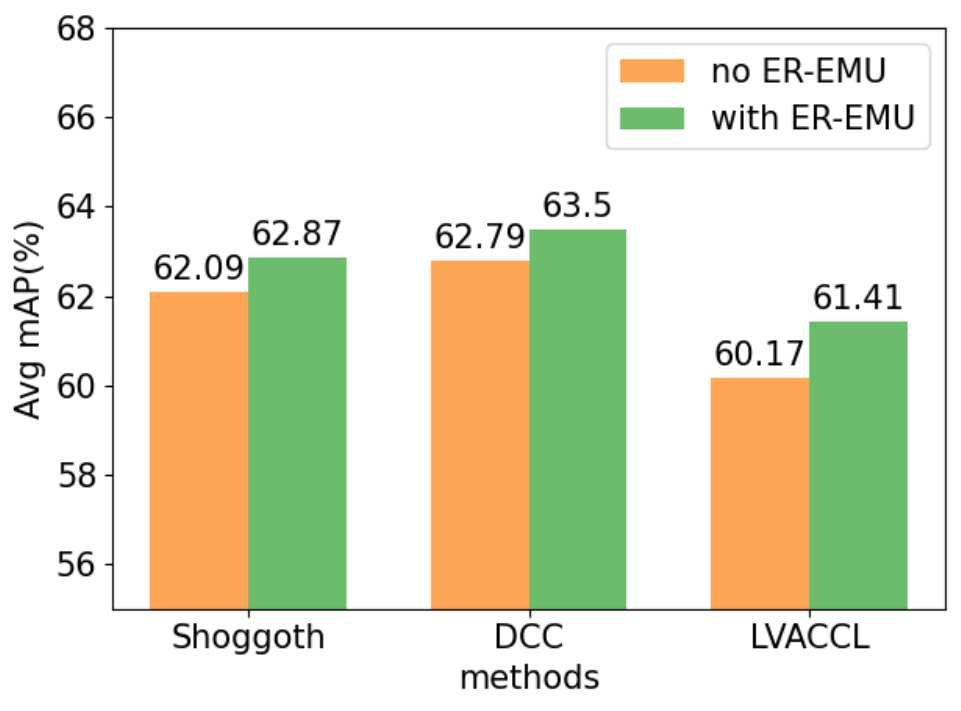}
    \caption{Time Span: 2D}
    \label{fig:sub2}
\end{subfigure}
\begin{subfigure}[b]{0.24\linewidth}
    \includegraphics[width=\linewidth]{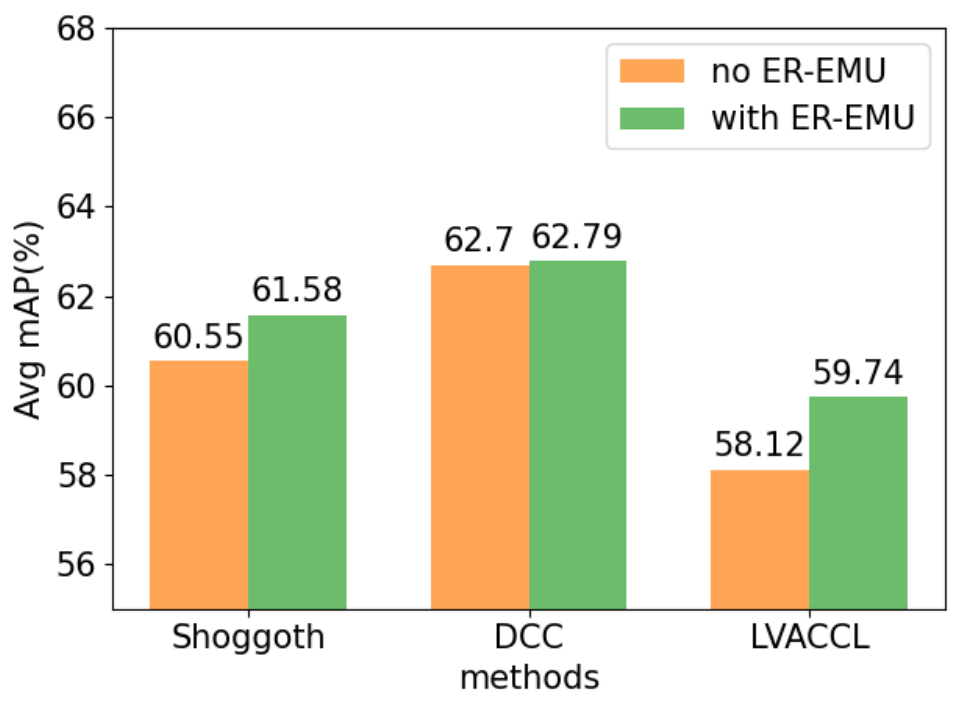}
    \caption{Time Span: 4D}
    \label{fig:sub3}
\end{subfigure}
\begin{subfigure}[b]{0.24\linewidth}
    \includegraphics[width=\linewidth]{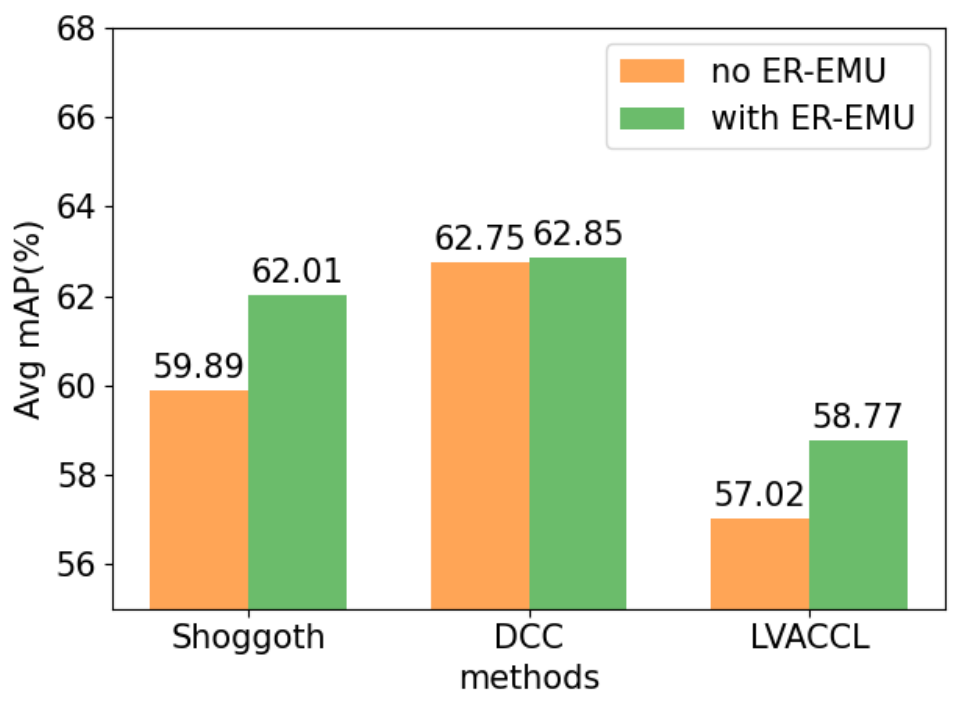}
    \caption{Time Span: 7D}
    \label{fig:sub4}
\end{subfigure}
\caption{Performance verification of ER-EMU.}
\label{fig:ER-EMU算法的性能验证}
\end{figure*}

\begin{figure*}[ht]
\centering

\begin{subfigure}[b]{0.25\linewidth}
    \includegraphics[width=\linewidth]{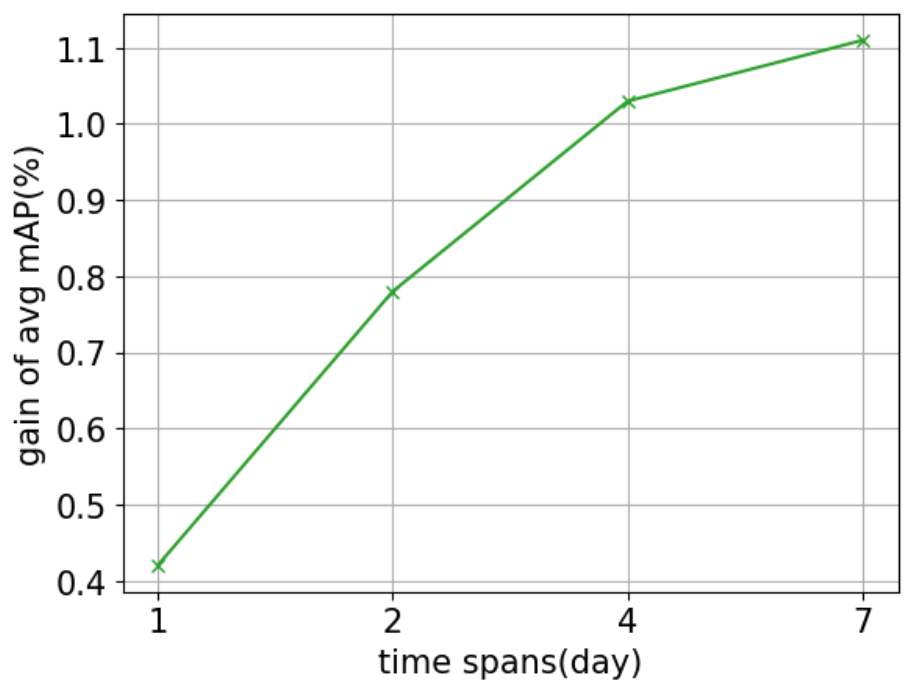}
    \caption{Shoggoth}
    \label{fig:sub1}
\end{subfigure}
\begin{subfigure}[b]{0.25\linewidth}
    \includegraphics[width=\linewidth]{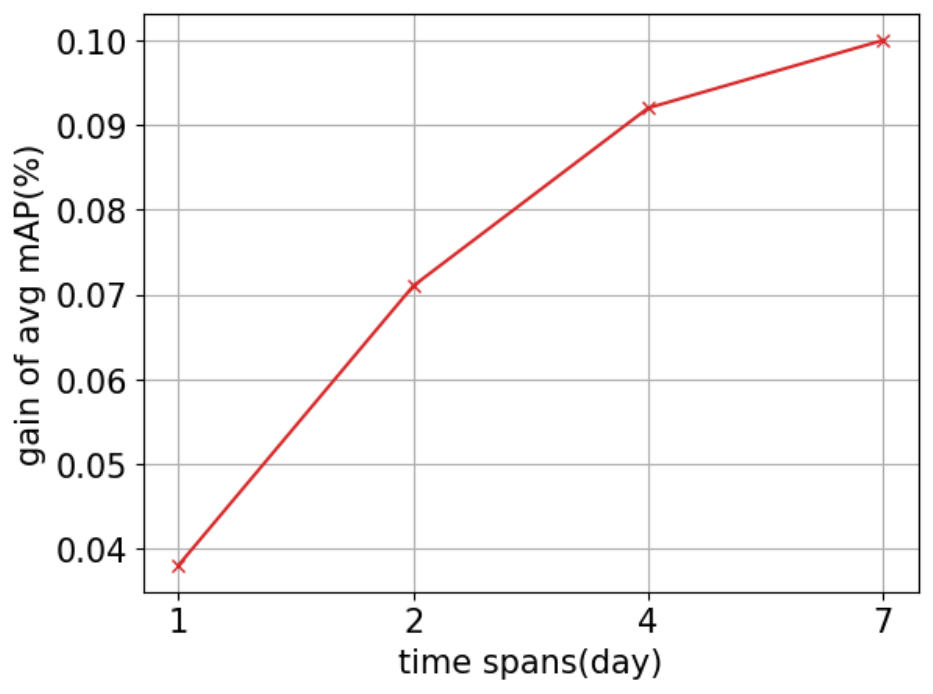}
    \caption{DCC}
    \label{fig:sub2}
\end{subfigure}
\begin{subfigure}[b]{0.25\linewidth}
    \includegraphics[width=\linewidth]{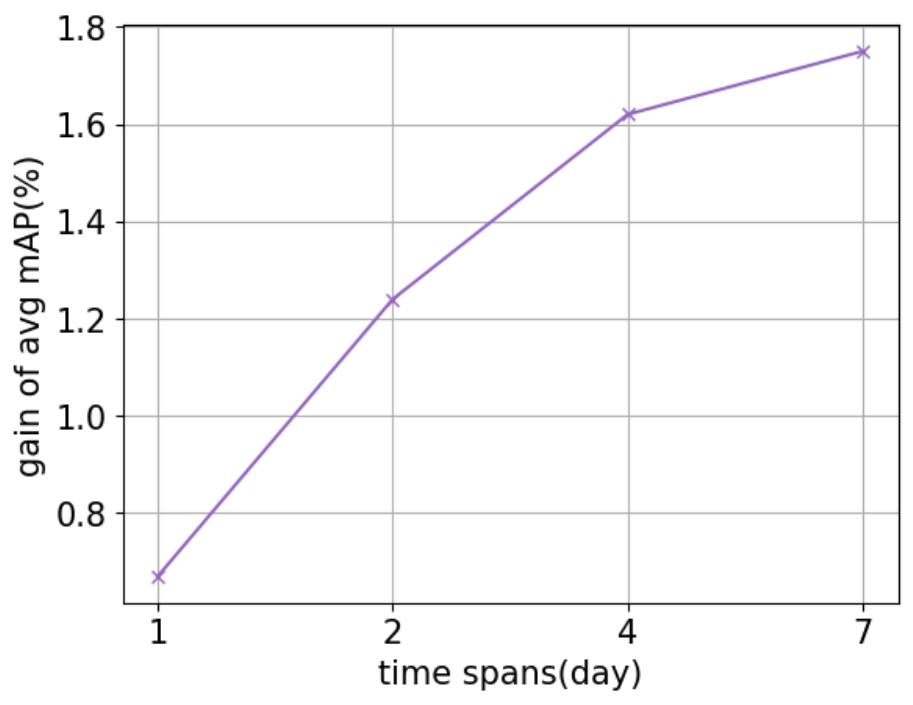}
    \caption{LVACCL}
    \label{fig:sub3}
\end{subfigure}
\caption{Gain variation of ER-EMU at different time span.}
\label{fig:ER-EMU算法在不同时间跨度上的增益变化}
\end{figure*}

\subsection{Ablation Study}

\begin{figure*}[ht]
\centering
\begin{subfigure}[b]{0.4\linewidth}
    \includegraphics[width=\linewidth]{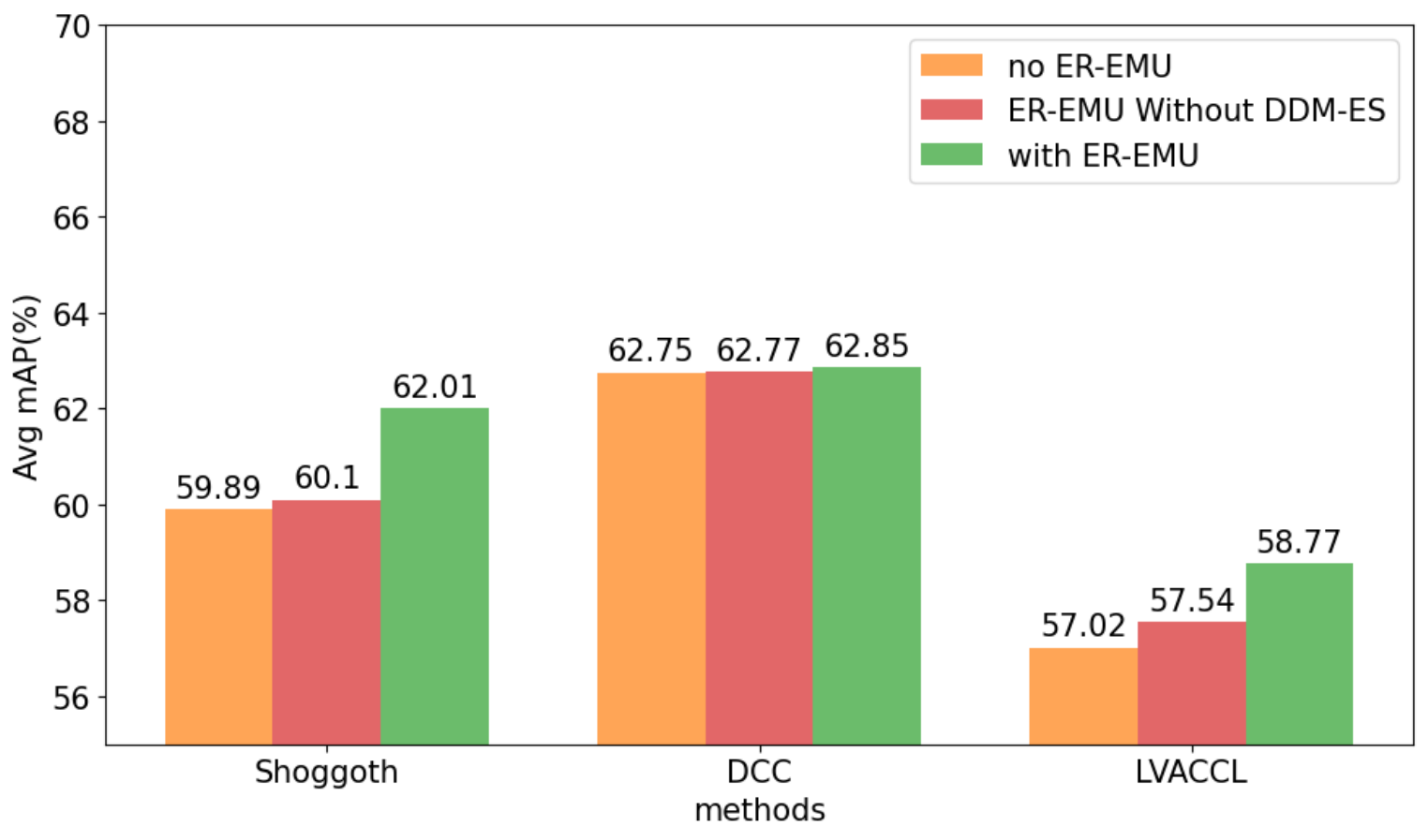}
    \caption{Experimental results of ablation study for ER-EMU.}
    \label{fig:ER-EMU算法的消融实验结果}
\end{subfigure}
\begin{subfigure}[b]{0.4\linewidth}
    \includegraphics[width=\linewidth]{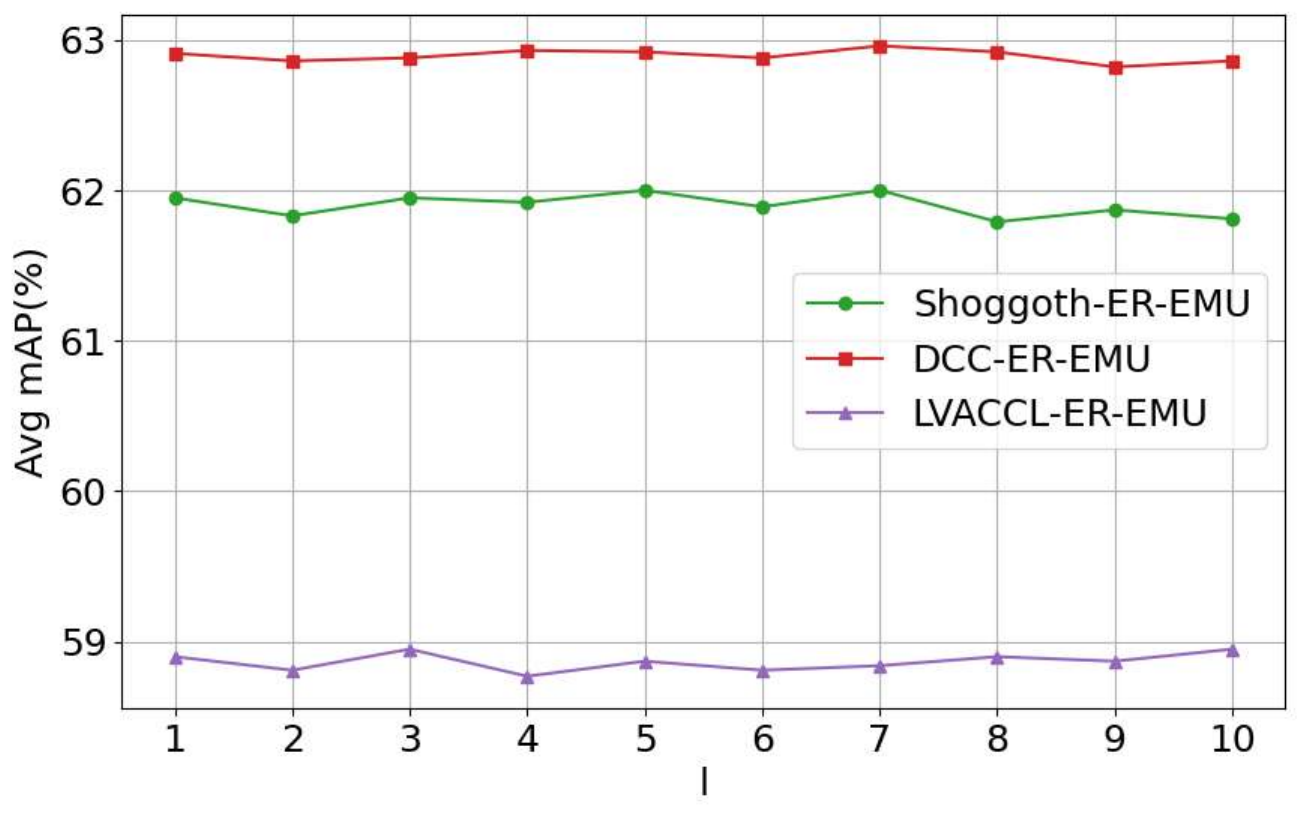}
    \caption{Parameter $l$ on the performance of ER-EMU.}
    \label{fig:经验回放超参数l}
\end{subfigure}

\caption{Gain variation of ER-EMU at different time span.}
\label{fig:ER-EMU算法在不同时间跨度上的增益变化}
\end{figure*}

To validate ER-EMU's edge model update performance, we applied the methods to 8 Bellevue camera video streams with varying time spans. Results in Figure \ref{fig:ER-EMU算法的性能验证} (1, 2, 4, 7 day spans) demonstrate ER-EMU enhances SOTA cloud-edge methods and exhibits robustness across time spans. This highlights the importance of leveraging past experiences and dynamically updating environmental models.

ER-EMU's smaller improvement on DCC is likely due to DCC's specific catastrophic forgetting optimizations and visual prompts. However, DCC still improved with ER-EMU, indicating ER-EMU's superiority.

Figure \ref{fig:ER-EMU算法在不同时间跨度上的增益变化} shows detection accuracy gains across time spans. Figures \ref{fig:ER-EMU算法在不同时间跨度上的增益变化} (Shoggoth, DCC, LVACCL) depict gain changes. While gains vary across methods, the pattern of increasing gain with longer time spans is similar, suggesting ER-EMU is more advantageous for longer streams. This relates to its ability to cache key historical samples and adaptively select data for retraining. ER-EMU achieves continuous gains in traffic environments with regular variations. However, due to limited model capacity, gains eventually slow down.

A core module of the ER-EMU algorithm lies in using the DDM-ES algorithm to select samples from the experience buffer for retraining the edge model in a domain distance-based manner. This section focuses on investigating the role of this algorithm within the overall ER-EMU algorithm.

To isolate ER-EMU's components, a trimmed ER-EMU (DDM-ES replaced with random sampling) and the full ER-EMU were applied to Shoggoth, DCC, and LVACCL using a week of video data from 8 Bellevue cameras. Figure \ref{fig:ER-EMU算法的消融实验结果} shows that the random sampling ER-EMU still improves performance, suggesting random historical data selection can alleviate catastrophic forgetting. The performance decrease of random sampling ER-EMU compared to the full ER-EMU is most significant for Shoggoth. This is likely because Shoggoth already uses random sampling for experience buffer updates, only lacking a priority-based selection mechanism; its update algorithm is similar to the trimmed ER-EMU. This, combined with the other methods' results, demonstrates the importance of the DDM-ES historical experience selection algorithm in ER-EMU.

\subsection{Influence of Parameter l on ER-EMU Algorithm Performance}

To investigate the influence of the hyperparameter $l$ in the ER-EMU algorithm, experiments were conducted with $l$ values ranging from 1 to 10, keeping all other parameters consistent.  Results, as depicted in Figure \ref{fig:经验回放超参数l}, demonstrate that variations in $l$ do not significantly impact the algorithm's performance. Therefore, the ER-EMU algorithm exhibits robustness and low sensitivity to the choice of the hyperparameter $l$.

\section{Conclusion}
This paper addresses the challenge of catastrophic forgetting in edge model retraining for cloud-edge collaborative object detection in traffic monitoring. It proposes ER-EMU, an edge model update algorithm based on adaptive experience replay. 
ER-EMU utilizes a limited-size experience buffer and a novel experience selection algorithm based on inter-domain distance metrics. This approach selects the most representative historical data, enriching the training set, improving generalization, and mitigating forgetting.  The experience selection leverages multi-kernel maximum mean discrepancy to quantify domain distance, prioritizing diverse historical data for replay. Additionally, a random sampling-based buffer update mechanism maintains data diversity and balance.
Experiments on the Bellevue traffic video dataset demonstrate ER-EMU's superiority over existing edge model update algorithms in various cloud-edge collaborative architectures. 

\bibliographystyle{splncs04}
\bibliography{refs}

\end{document}